# Study of Some Recent Crossovers Effects on Speed and Accuracy of Genetic Algorithm, Using Symmetric Travelling Salesman Problem

Hassan Ismkhan
Instructor at University of Bonab (Binab)
Computer Engineering Department, University of Bonab (Binab), East Azerbaijan, Iran
H.Ismkhan@bonabu.ac.ir

Kamran Zamanifar
Associate Professor at University of Isfahan
Computer Engineering Department, University of Isfahan, Isfahan, Iran
zamanifar@eng.ui.ac.ir

## ABSTRACT
The Travelling Salesman Problem (TSP) is one of the most famous optimization problems. The Genetic Algorithm (GA) is one of metaheuristics that have been applied to TSP. The Crossover and mutation operators are two important elements of GA. There are many TSP solver crossover operators. In this paper, we state implementation of some recent TSP solver crossovers at first and then we use each of them in GA to solve some Symmetric TSP (STSP) instances and finally compare their effects on speed and accuracy of presented GA.

## 1. INTRODUCTION
After introducing Genetic Algorithm (GA) by Holand [1] many crossover operators have been invented by researchers because the performance of GA depends on an ability of these operators. PMX [1] is one of first crossovers proposed by Goldberg and Lingle in 1985. Reference [12] stated some important shortcomings of PMX and to overcome them proposed extended PMX (EPMX). DPX [9][10] is another crossover that produces child with greedy reconnect of common edges in two parents. References [9][10] use DPX in their Genetic Local Search (GLS) algorithms. Greedy Subtour Crossovers (GSXs) [7][8][9] family are another groups of crossovers that operate fast. GSX-2 [8] is improved version of GSX-0 [6] and GSX-1 [7].

In this paper, we compare some of these crossovers on speed and accuracy so the rest of paper organized as follows: we represent some crossovers in section 2. In section 3, we show implementation of our GA. In section 4 we put forward our experimental results and compare stated crossovers and finally we summarize paper in section 5.

## 2. REPRESENTATION OF SOME RECENT CROSSOVERS
In this section, we represent some recent GA crossovers. In our example we use the a graph with eight nodes {1,2,3,4,5,6,7,8} that its edges weight are as Figure 1.

### 2.1 EPMX Crossover
Partially Mapped Crossover (PMX) is one of first genetic operator. It produces two children from two parents by exchanging nodes between two arbitrary points.

|   | 1 | 2 | 3 | 4 | 5 | 6 | 7 | 8 |
|---|---|---|---|---|---|---|---|---|
| 1 | 0 | 12 | 19 | 31 | 22 | 17 | 23 | 12 |
| 2 | 12 | 0 | 15 | 37 | 21 | 28 | 35 | 22 |
| 3 | 19 | 15 | 0 | 50 | 36 | 35 | 35 | 21 |
| 4 | 31 | 37 | 50 | 0 | 20 | 21 | 37 | 38 |
| 5 | 22 | 21 | 36 | 20 | 0 | 25 | 40 | 33 |
| 6 | 17 | 28 | 35 | 21 | 25 | 0 | 16 | 18 |
| 7 | 23 | 35 | 35 | 37 | 40 | 16 | 0 | 14 |
| 8 | 12 | 22 | 21 | 38 | 33 | 18 | 14 | 0 |

Figure. 1 Graph edges weight

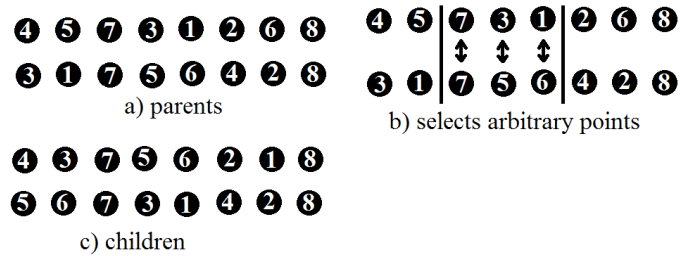

Figure.2 PMX example

In PMX example, node 7 is participated in two mapped areas but PMX is unable to detect same nodes from mapped areas and also is double point crossover and is not suitable to solve TSP and in many cases maybe produces repetitive children [12]. Reference [12] tried to overcome these shortcomings and proposed Extended PMX (EPMX). It selects one arbitrary point and exchanges unique nodes before these arbitrary points and produce two children.

For example of EPMX given father=1-2-3-4-5-6-7-8 and mother=1-4-8-6-2-3-5-7. Suppose that Arbitrary Point=4 so father=1-2-3-4|5-6-7-8 and mother=1-4-8-6|2-3-5-7 are divided to two sub-lists. 2 and 3 from first sub-list of father are not repeated in first sub-list of mother and 8 and 6 from first sub-list of mother are not repeated in first sub-list of father so {(2↔6),(3↔8)} formed exchanges so children produces as child1=1-4-8-6-5-3-7-2, child2=1-2-3-4-8-6-5-7.

## 2.2 Greedy Crossover (GX)

GXs select a node and copy it to child then it probes witch of its neighbors is nearest to it, so the nearest one is copied to child, and this process is continued until child tour be completed. We show some previous versions of GX by example in figure 3. In this example we use a graph with 8 nodes that its edges cost are as distance matrix in figure 1.

| father: | 4 | 5 | 7 | 3 | 2 | 1 | 6 | 8 |
|---|---|---|---|---|---|---|---|---|
| mother: | 5 | 1 | 7 | 3 | 6 | 2 | 4 | 8 |

**step 1:**
First node, is selected randomly and copied to tour. References [3][4] always use same node at start. Please suppose that selected number would be 1.

| child: | 1 | | | | | | | |
|---|---|---|---|---|---|---|---|---|

**step 2:**

| father: | 4 | 5 | 7 | 3 | 2 | 1 | 6 | 8 |
|---|---|---|---|---|---|---|---|---|
| mother: | 5 | 1 | 7 | 3 | 6 | 2 | 4 | 8 |

In each step, four neighbors of recent selected node are considered and which is closer to it is selected.
2, 6, 5 and 7 are neighbors of 1 and 2 is closer to it so is copied to child.

| child: | 1 | 2 | | | | | | |
|---|---|---|---|---|---|---|---|---|

Step 2 repeated until tour is be completed.

**Special cases**

| father: | 4 | 5 | 7 | 3 | 2 | 1 | 6 | 8 |
|---|---|---|---|---|---|---|---|---|
| mother: | 5 | 1 | 7 | 3 | 6 | 2 | 4 | 8 |

3, 1, 6, and 4 are neighbors of 2 and 1 are closer to it but 1 is already exist in child then we cannot copy it to child.

| child: | 1 | 2 | ? | | | | | |
|---|---|---|---|---|---|---|---|---|

In this cases:
- GX [2] selects next node randomly.
- GX [3][4][5][6] considers another three nodes.
- If all of four nodes would be in child then:
  - ➢ References [3][4] select next node randomly.
  - ➢ Reference [5] chooses closer node (to recent selected node) among 20 random remaining nodes that are not copied to child yet.
  - ➢ Reference [6] operates very greedy and selects closer node among all remaining nodes. This version has been named very greedy crossover (VGX).

Figure 3. GXs review

## 2.3 Unnamed Heuristic Crossover

Presented crossover in [14] is unnamed and operates heuristically so here we name it UHX (unnamed heuristic crossover). UHX starts with random current city and copies it to child. It puts four pointers to right and left neighbors of current city in both parents and then compares which pointed city is nearest to current city and has not been copied to child yet. Such a node is selected as current city, copied to child and its pointer goes forward in its direction (for example if the father right pointer's city is nearest then right pointer in father goes one forward in right direction). Figure 4 shows UHX example.

## 2.4 Improved Greedy Subtour Crossover (GSX-2)

GSX-2 [9] is improved version of GSX-0 [7] and GSX-1 [8]. GSX-0 is first version of GSX family. It selects a node randomly at first. In second step, nodes in right side of father and nodes in left side of mother are copied to child. This process is continued until an existing node in child is met. At last step, the remaining nodes are copied to random places of child. We show this crossover by example in figure 5.

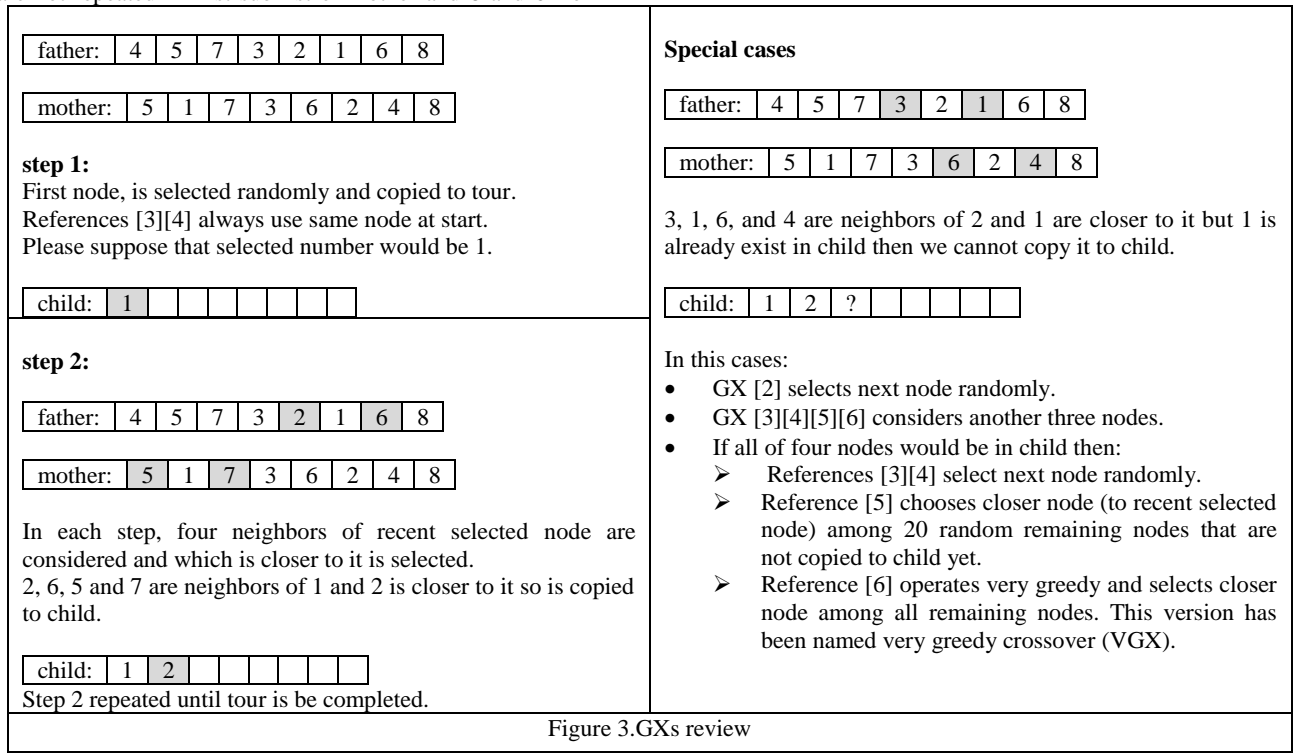

Figure 4. UHX

In Fig.5 after that node 5, has been already included in child, is met, GSX-0 fill remaining places with random nodes but GSX-1 fill remaining nodes in order of one of parents. In some cases, GSX-1 has shortcoming and produces repetitive tour. Fig.6 shows this problem by example.

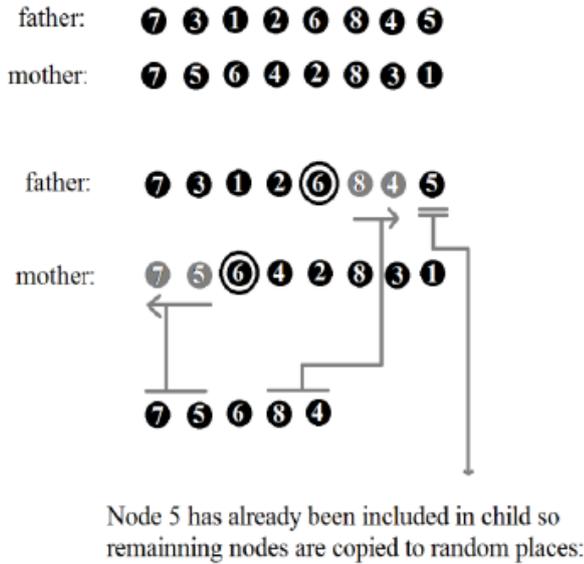

Figure 5. GSX-0

To overcome above shortcoming reference [9] proposes GSX-2. GSX-2 to decide which direction should be selected probes left and right nodes around first random selected node. In above example right node of 4 in father is 6 and left node of 4 in mother is also 6 so GSX-2 selects right direction for mother.

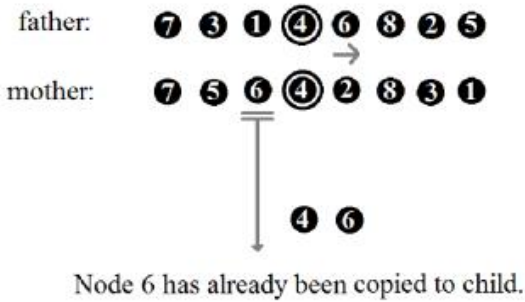

Figure 6. GSX-1 shortcoming

## 2.5 Distance Preserving Operator (DPX)

DPX[9][10] operates as follows: it detects common subpathes of two parents at first then reconnects them greedy and produces child. Figure 7 is DPX example that uses presented edges weight of graph in figure 1.

## 3. GA REPRESANTATION

We present GA as figure 8 pseudo-codes. This algorithm initializes population randomly. Lines 2 to 8 produce children by crossover and mutates them by applying our local searches: 2opt_move_based_LS and 3opt_move_based_LS [13].

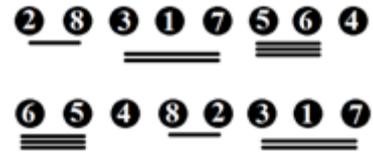

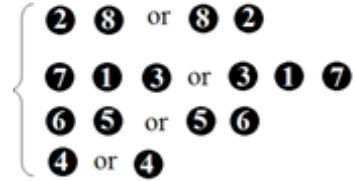

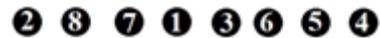

Figure.7 DPX example

If one of produced children is better than one of population individuals then "while" loop in line 2 will be continued.

1) initialize population with random tours
2) **while** population is changed
3)     **for** i=1 to Generation-Size
4)         Select father and mother from population
5)         child ← operate one of GX versions on father and mother
6)         operate 2_Opt_move_based LS on child
7)         operate 3_Opt_move_based LS on child
8)         add child to population
9)     reduce population
10) return best individual of population

Figure.8 GA pseudo-codes

## 4. Experiments

We implemented all of algorithms with language c# and used .NET 2008. We ran all experiments on AMD Dual Core 2.6 GHZ. We tested each of stated crossovers in mentioned GA on following seven instances: eil51, eil76, eil101, kroA100, kroA200, a280 and lin318 which are all from TSPLIB[11]. We set population size and generation size 50 and 500 respectively. For each instance, 10 runs were performed.

Table I, shows results of experiments. In this table "Best length", "Average length" and "Worst length" show the best, average, and worst tour lengths of twenty runs, respectively. "Number of repeat "while" loop in lines 2 to 8" column points out how many times lines 2 to 8 in figure 8 is executed also "average time" column gives the average convergence time of GA per each crossover in second. We summarize fourth and seventh columns in figures 9 and 10 respectively. In "Best length", "Average length" and "Worst length" columns the values in parentheses is result of calculating

$$\frac{\text{cost of solution found} - \text{known optimum cost}}{\text{known optimum cost}} \times 100$$

These results show that when GA uses heuristic crossovers (as GXs, UHX and DPX) has more accuracy than when uses non-heuristic crossovers (as PMX, EPMX, GSX-2). Sixth column shows that when GA uses non heuristic crossovers repeated lines 2 to 8 more than when uses heuristic crossovers because non heuristic crossovers have more diversity than heuristic crossovers and they can produce variety of different children then they are delaying GA.

Table. 1: Results

| Problem name | Crossover name | Best length (quality) | Average length (quality) | Worst length (quality) | Number of repeat "while" loop in lines 2 to 8 | convergence time (second) |
|---|---|---|---|---|---|---|
| Eil51 | PMX | 434(1.88) | 444.8(4.41) | 451(5.87) | 52 | 3.39144 |
| | EPMX | 433(1.64) | 445(4.46) | 459(7.75) | 53 | 3.43044 |
| | GSX-2 | 428(0.47) | 446.1(4.72) | 468(9.86) | 42 | 1.56 |
| | GX[2] | 716(68.08) | 747.7(75.52) | 792(85.92) | 30 | 3.64728 |
| | GX[3][4] | 430(0.94) | 443.8(4.18) | 456(7.04) | 32 | 5.29308 |
| | GX[5] | 426(0) | 432(1.41) | 437(2.58) | 27 | 4.72992 |
| | VGX | 430(0.94) | 431.5(1.29) | 434(1.88) | 24 | 5.31492 |
| | UHX | 426(0) | 430.5(1.06) | 438(2.82) | 29 | 4.23228 |
| | DPX | 429(0.7) | 431.5(1.29) | 434(1.88) | 21 | 1.51632 |
| Eil76 | PMX | 568(5.58) | 580.7(7.94) | 596(10.78) | 89 | 7.16352 |
| | EPMX | 554(2.97) | 576.2(7.1) | 603(12.08) | 91 | 7.29924 |
| | GSX-2 | 557(3.53) | 576.8(7.21) | 583(8.36) | 68 | 3.159 |
| | GX[2] | 1083(101.3) | 1141.2(112.12) | 1189(121) | 42 | 7.2696 |
| | GX[3][4] | 558(3.72) | 570.5(6.04) | 579(7.62) | 41 | 9.98712 |
| | GX[5] | 548(1.86) | 555.9(3.33) | 569(5.76) | 46 | 11.38644 |
| | VGX | 543(0.93) | 546.9(1.65) | 555(3.16) | 36 | 10.3506 |
| | UHX | 542(0.74) | 549.6(2.16) | 557(3.53) | 40 | 8.84988 |
| | DPX | 546(1.49) | 550.7(2.36) | 553(2.79) | 21 | 2.38524 |
| kroA100 | PMX | 21598(1.48) | 22884(7.53) | 24368(14.5) | 117 | 11.63604 |
| | EPMX | 22295(4.76) | 22959.6(7.88) | 24013(12.83) | 119 | 11.80764 |
| | GSX-2 | 21940(3.09) | 22492.5(5.69) | 23068(8.39) | 105 | 5.91396 |
| | GX[2] | 68300(220.93) | 71045.2(233.83) | 73942(247.44) | 37 | 8.53944 |
| | GX[3][4] | 21639(1.68) | 22986.3(8.01) | 24805(16.55) | 46 | 14.85432 |
| | GX[5] | 21671(1.83) | 22036.1(3.54) | 22821(7.23) | 68 | 22.29396 |
| | VGX | 21320(0.18) | 21491.8(0.99) | 21706(1.99) | 45 | 15.21156 |
| | UHX | 21320(0.18) | 21440.4(0.74) | 21573(1.37) | 42 | 15.08676 |
| | DPX | 21393(0.52) | 21743.8(2.17) | 23181(8.92) | 34 | 5.031 |
| kroA200 | PMX | 31379(6.85) | 32745.3(11.5) | 34068(16) | 262 | 43.85472 |
| | EPMX | 32347(10.14) | 33264.8(13.27) | 34297(16.78) | 262 | 43.46472 |
| | GSX-2 | 31378(6.84) | 32437.8(10.45) | 33440(13.87) | 243 | 23.37192 |
| | GX[2] | 150692(413.12) | 153051.8(421.15) | 156852(434.09) | 40 | 17.68416 |
| | GX[3][4] | 31351(6.75) | 35171.5(19.76) | 39389(34.12) | 120 | 75.41352 |
| | GX[5] | 30335(3.29) | 31141.9(6.04) | 32314(10.03) | 113 | 73.02828 |
| | VGX | 29706(1.15) | 29995.6(2.14) | 30392(3.49) | 57 | 56.73096 |
| | UHX | 29680(1.06) | 29950.6(1.98) | 30872(5.12) | 81 | 38.35104 |
| | DPX | 30079(2.42) | 30532.2(3.96) | 31077(5.82) | 47 | 15.57192 |
| A280 | PMX | 2916(13.07) | 3067.2(18.93) | 3222(24.93) | 363 | 76.22316 |
| | EPMX | 2887(11.94) | 3081(19.46) | 3169(22.88) | 380 | 79.69104 |
| | GSX-2 | 2923(13.34) | 3002.5(16.42) | 3066(18.88) | 364 | 44.57076 |
| | GX[2] | 15901(516.56) | 16180.5(527.39) | 16625(544.63) | 37 | 20.92428 |
| | GX[3][4] | 3328(29.04) | 3645.5(41.35) | 4202(62.93) | 125 | 102.33288 |
| | GX[5] | 2803(8.69) | 2898.9(12.4) | 3006(16.56) | 162 | 132.8574 |
| | VGX | 2639(2.33) | 2662.4(3.23) | 2683(4.03) | 69 | 62.65116 |
| | UHX | 2649(2.71) | 2693.9(4.46) | 2766(7.25) | 80 | 61.1598 |
| | DPX | 2651(2.79) | 2720.4(5.48) | 2776(7.64) | 49 | 24.33756 |
| Lin318 | PMX | 46073(9.62) | 48673.1(15.81) | 50426(19.98) | 444 | 107.37792 |
| | EPMX | 46956(11.72) | 48373.9(15.1) | 50058(19.1) | 465 | 112.4214 |
| | GSX-2 | 45971(9.38) | 47307.6(12.56) | 48573(15.57) | 440 | 62.4078 |
| | GX[2] | 269330(540.82) | 277900.6(561.21) | 282884(573.07) | 42 | 28.6884 |
| | GX[3][4] | 51425(22.36) | 56761.3(35.05) | 65868(56.72) | 190 | 185.20788 |
| | GX[5] | 45208(7.56) | 45864.8(9.13) | 46961(11.73) | 212 | 208.0182 |
| | VGX | 43293(3.01) | 43756.6(4.11) | 44327(5.47) | 78 | 137.82756 |
| | UHX | 43354(3.15) | 44003.7(4.7) | 45078(7.25) | 126 | 81.6816 |
| | DPX | 44381(5.6) | 45052.9(7.19) | 45814(9.01) | 68 | 40.86732 |

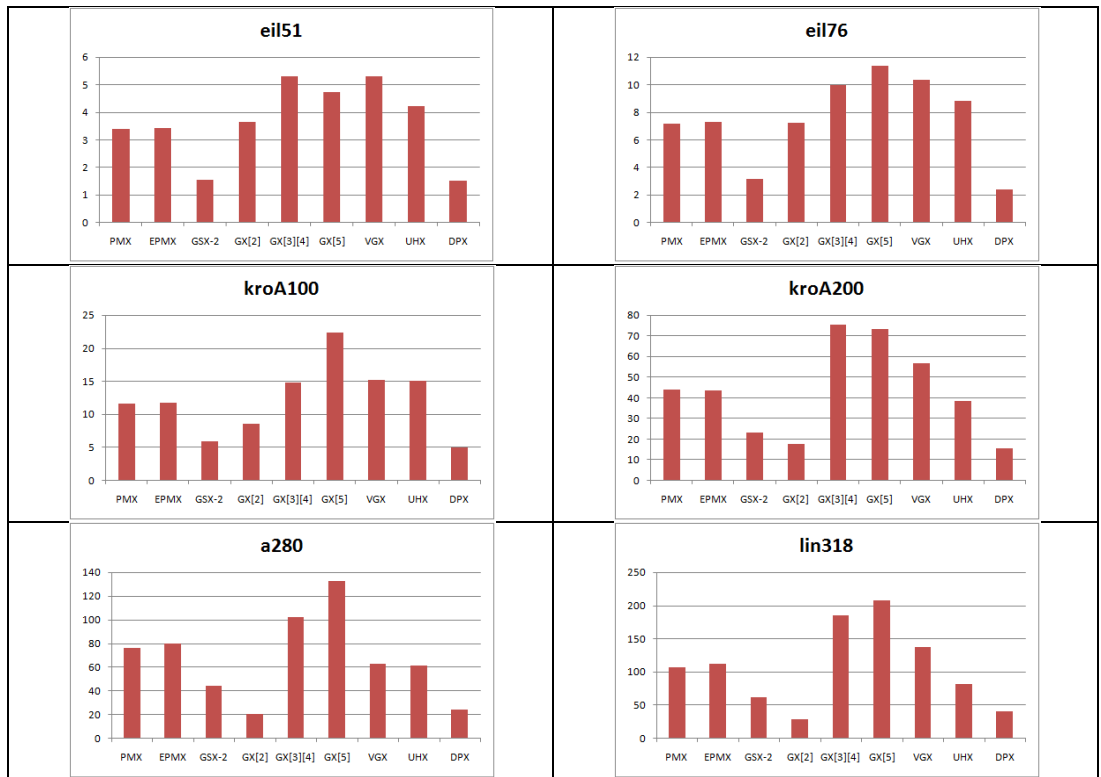

Figure.9 Average length of problems per each of crossovers

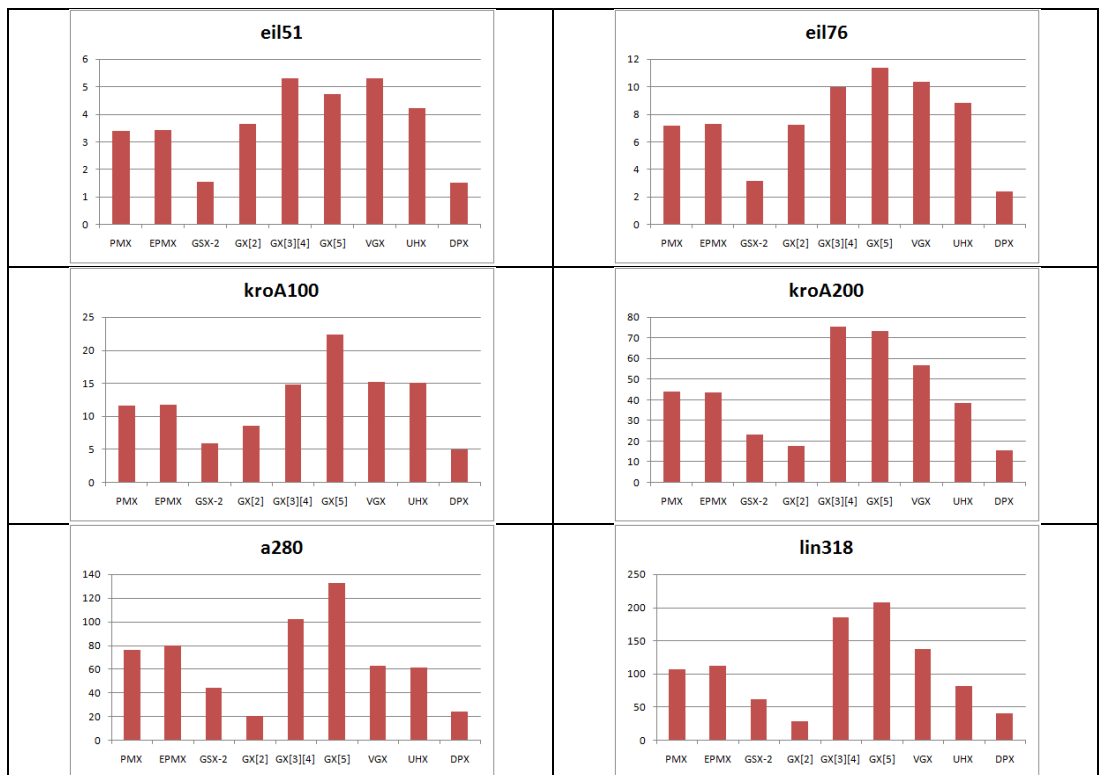

Figure.10 Average time of GA convergence when uses each of crossovers

## 5. CONCLUSION

In this paper, we represented some recent crossovers and exhibited implementation of our GA. We compared crossovers on speed and accuracy by using and testing them in GA. We implemented all of algorithms in C# and used .NET framework in our experiments. Our experiments shown that heuristic crossovers have more accuracy than other non-heuristic one and non-heuristic crossovers same as GSX-2 have more diversity and they can produce variety of different children.